\documentclass[10pt,twocolumn,letterpaper]{article}

\usepackage{cvpr}
\usepackage{times}
\usepackage{epsfig}
\usepackage{graphicx}
\usepackage{amsmath}
\usepackage{amssymb}
\usepackage{paralist}
\usepackage{algorithm}
\usepackage{algorithmic}
\usepackage{booktabs}
\usepackage{multirow}
\usepackage{mathtools}

\newcommand{\abc}[1]{\textcolor{black}{#1}}
\newcommand{\abcn}[1]{\textcolor{black}{#1}}
\newcommand{\abcnn}[1]{\textcolor{black}{#1}}
\newcommand{\nabc}[1]{\textcolor{black}{#1}}
\newcommand{\nnabc}[1]{\textcolor{black}{#1}}

\newcommand{\yty}[1]{\textcolor{black}{#1}}
\newcommand{\nyty}[1]{\textcolor{black}{#1}}
\newcommand{\ytyn}[1]{\textcolor{black}{#1}}

\newcommand{\s}[1]{^{(#1)}}
\newcommand{\refeqn}[1]{(\ref{#1})}
\newcommand{\NOTE}[1]{\textcolor{red}{[NOTE: #1]}}
\newcommand{\CUT}[1]{}
\newcommand{\first}[1]{\textcolor{red}{#1}}
\newcommand{\second}[1]{\textcolor{green}{#1}}
\newcommand{\third}[1]{\textcolor{blue}{#1}}
\newcommand{\tabincell}[2]{\begin{tabular}{@{}#1@{}}#2\end{tabular}}  
\DeclarePairedDelimiter\ceil{\lceil}{\rceil}

\usepackage[font=footnotesize]{caption}
\usepackage[square,numbers,sort&compress]{natbib}

\usepackage[pagebackref=true,breaklinks=true,letterpaper=true,colorlinks,bookmarks=false]{hyperref}

\cvprfinalcopy 

\def\cvprPaperID{4243} 

\ifcvprfinal\fi
\begin{document}

\title{ROAM: Recurrently Optimizing Tracking Model}

\author{Tianyu Yang$^{1,2}$ \quad Pengfei Xu$^3$ \quad Runbo Hu$^3$ \quad Hua Chai$^3$  \quad Antoni B. Chan$^2$\\
	$^1$ Tencent AI Lab \quad $^2$ City University of Hong Kong  \quad $^3$ Didi Chuxing \\
}

\maketitle

\begin{abstract}
	In this paper, we design a tracking model consisting of response generation and bounding box regression, where the first component produces a heat map to indicate the presence of the object at different positions and the second part regresses the relative bounding box shifts to anchors mounted on sliding-window locations. Thanks to the resizable convolutional filters used in both components to adapt to the shape changes of objects,  our tracking model does not need to enumerate different sized anchors, thus saving model parameters.
	To effectively adapt the model to appearance variations, we propose to offline train a recurrent neural optimizer to update tracking model in a meta-learning setting,  which can converge the model in a few gradient steps. This improves the convergence speed of updating the tracking model while achieving better performance. We extensively evaluate our trackers, ROAM and ROAM++, on the OTB, VOT, LaSOT, GOT-10K and TrackingNet benchmark and our methods perform favorably against state-of-the-art algorithms.
\end{abstract}

\section{Introduction}
Generic visual object tracking is the task of estimating the bounding box of a target in a video sequence given only its initial position. Typically, the preliminary model learned from the first frame needs to be updated continuously to adapt to the target's appearance variations caused by rotation, illumination, occlusion, deformation, etc.  However, it is challenging to optimize the initial learned model efficiently and effectively as tracking proceeds.  Training samples for model updating are usually collected based on estimated bounding boxes, which could be inaccurate. Those small errors will accumulate over time, gradually resulting in model degradation.

To avoid model updating, which may introduce unreliable training samples that ruin the model, several approaches \cite{Bertinetto2016, Tao2016} investigate tracking by only comparing the first frame with the subsequent frames, using a similarity function based on a learned discriminant and invariant deep Siamese feature embedding. However, training such a deep representation is difficult due to drastic appearance variations that commonly emerge in long-term tracking. Other methods either update the model via an exponential moving average of templates \cite{Henriques2015, Valmadre2017}, which marginally improves the performance, or optimize the model with hand-designed SGD methods \cite{Nam2016, Song2017}, which needs numerous iterations to converge \abc{thus preventing real-time speed}. 
\abcn{Limiting the number of SGD iterations can allow near real-time speeds, but at the expense of poor quality model updates due to the loss function not being optimized sufficiently.}

In recent years, much effort has been done on localizing the object using robust online learned classifier, while few attention is paid on designing accurate bounding box estimation. Most trackers simply resort to multi-scale search by assuming that the object aspect ratio does not change during tracking, which is often violated in real world. Recently, SiamRPN \cite{Li2018} borrows the idea of region proposal networks \cite{Ren2015} in object detection to decompose tracking task into two branches: 1) classifying the target from the background, and 2) regressing the accurate bounding box based with reference to anchor boxes mounted on different positions. 
\nyty{As is shown on the VOT benchmarks \cite{Kristan2016, Kristan2017},} \nabc{SiamRPN} achieves higher precision on bounding box estimation but suffers lower robustness 
compared with state-of-the-art methods \cite{Lu2018, Danelljan2016, Danelljan2016-1} 
due to no online model updating. \nyty{Furthermore, SiamRPN mounts anchors with different aspect ratios on every spatial location of the feature map to handle possible shape changes, which is 
redundant in both computation and storage. }

In this paper, we propose a tracking framework which is composed of two modules: response generation and bounding box regression, where the first component produces a response map to indicate the possibility of covering the object for anchor boxes mounted on sliding-window positions, and the second part predicts bounding box shifts from the anchors to get refined rectangles. Instead of enumerating different aspect ratios of anchors as in SiamRPN, we propose to use only one sized anchor for each position and adapt it to shape changes by resizing its corresponding convolutional filter using bilinear interpolation, which saves model parameters and computing time.
To effectively adapt the tracking model to appearance changes during tracking, we propose a recurrent model optimization method to learn a more effective gradient descent that converges the model update in 1-2 steps, and generalizes better to future frames. 
The key idea is to train a neural optimizer that can converge the tracking model to a good solution in a few gradient steps. During the training phase, the tracking model is first updated using the neural optimizer, and then it is applied on future frames to obtain an error signal for minimization. Under this particular setting, the resulting optimizer converges the tracking classifier significant faster than SGD-based optimizers, especially for \abcn{learning the initial tracking model.}
In summary, our contributions are:
\begin{compactitem}
	\item We propose a tracking model consisting of resizable response generator and bounding box regressor, where only one sized anchor is used on each spatial position and its corresponding convolutional filter could be adapted to shape variations by bilinear interpolation.
	\item We propose a recurrent neural optimizer, which is trained in a meta-learning setting, that recurrently updates the tracking model with faster convergence.
	\item We conduct comprehensive experiments on large scale datasets including OTB, VOT, LaSOT, GOT10k and TrackingNet, and our trackers achieve favorable performance compared with the state-of-the-art.
\end{compactitem}


\section{Related Work}
\textbf{Visual Tracking}
Predicting a heat map to indicate the position of object is commonly used in visual tracking community  \cite{Henriques2015, Danelljan2016, Bertinetto2016-1, Bertinetto2016, Galoogahi2015, Bertinetto2016, Wang2019}. Among them, SiamFC  \cite{Bertinetto2016} is one of the most popular methods due to its fast speed and good performance.  However,  most response generation based trackers, including SiamFC, estimate the bounding box via simple multi-scale search mechanism, which is unable to handle aspect ratio changes. To addresses this issue, recent SiamRPN \cite{Li2018} and its extensions \cite{Li2019, Fan2019, Zhang2019} propose to train a bounding box regressor as in object detection \cite{Ren2015}, showing impressive performance.  Different from these algorithms which enumerates a set of predefined anchors with different aspect ratios on each spatial position, we adopt a resizable anchor to adapt the shape variation of object, which saves model parameters and computing time. 
 
Online model updating is another important module that SiamFC lacks. Recent works improve SiamFC \cite{Bertinetto2016} by introducing various model updating strategies, including recurrent generation of target template filters through a convolutional LSTM  \cite{Yang2017}, a dynamic memory network \cite{Yang2018, Yang2019}, where object information is written into and read from an addressable external memory, and distractor-aware incremental learning \cite{Zhu2018}, which make use of hard-negative templates around the target to suppress distractors. It should be noted that all these algorithms essentially achieve model updating  by linearly interpolating old target templates with the newly generated one, in which the major difference is how to control the weights when combining them. This is far from optimal compared with optimization methods using gradient decent,  which minimize the tracking loss directly to adapt to new target appearances. 
Instead of using a Siamese network to build the convolutional filter, other methods \cite{Song2017, Lu2018, Park2018} generate the filter by performing gradient decent on the first frame, which could be continuously optimized during subsequent frames. In particular, \cite{Park2018} proposes to train the {\em initial} tracking model in a meta-learning setting, which shows promising results. However, it still uses traditional SGD  to optimize the tracking model during the subsequent frames, which is not effective to adapt to new appearance and slow in updating the model. In contrast to these trackers, \nyty{our off-line learned recurrent neural optimizer applies meta-learning on both the initial model and model updates, which allows model initialization and updating in only one or two gradient steps}, resulting in much faster runtime speed, and better accuracy. 

\textbf{Learning to Learn.} Learning to learn or meta-learning has a long history \cite{schmidhuber1987evolutionary, bengio1992optimization, naik1992meta}. With the recent successes of applying meta-learning on few-shot classification \cite{Ravi2017, Munkhdalai2017} and reinforcement learning  \cite{Finn2017, Schulman2018}, it has regained attention. The pioneering work  \cite{Andrychowicz2016} 
designs an off-line learned optimizer using gradient decent and shows promising performance compared with traditional optimization methods. However, it does not generalize well for large numbers of descent step. To mitigate this problem,  \cite{Lv2017} proposes several training techniques,  including parameters scaling and combination with convex functions to coordinate the learning process of the optimizer. \cite{Wichrowska2017} also  addresses 
this issue by designing a hierarchical RNN architecture with dynamically adapted input and output scaling. In contrast to other works that output an increment for each parameter update, which is prone to overfitting due to different gradient scales, we instead associate an adaptive learning rate produced by a recurrent neural network with the computed gradient for fast convergence of the model update.


\begin{figure*}
	\centering
	\includegraphics[width=0.8\linewidth]{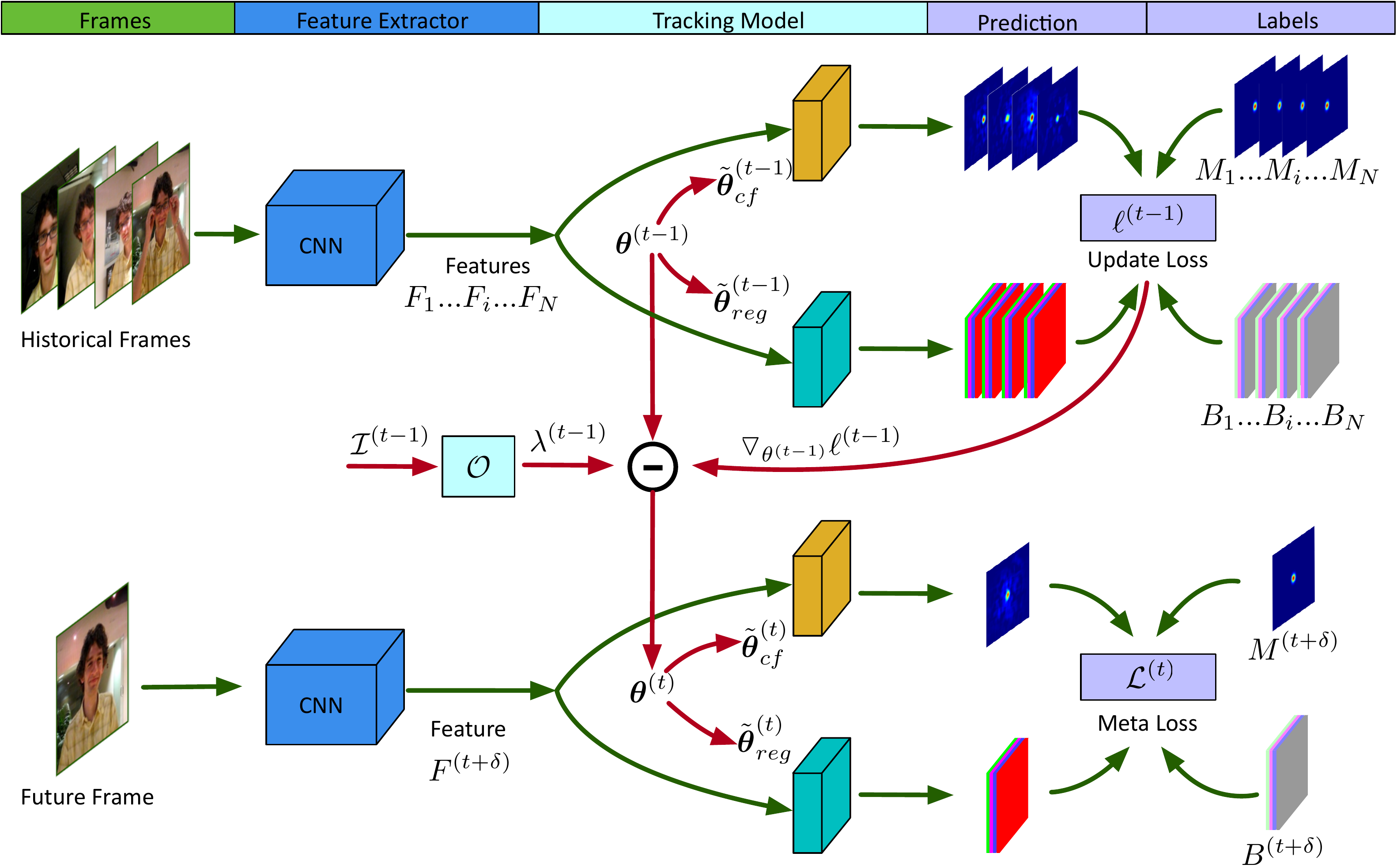}
		\vspace{-2ex}
	\caption{Pipeline of ROAM++. Given a mini-batch of training patches, \ytyn{which are cropped based on the predicted object boxes}, deep features are extracted by \emph{Feature Extractor}.
	\nnabc{The fixed-size \emph{Tracking Model} $\theta^{(t-1)}$ is warped to the current target size yielding the warped tracking model $\tilde{\theta}^{(t-1)}$, as in (\ref{eq:00}, \ref{eq:0}).}
	 The response map and bounding boxes are then predicted for each sample using $\tilde{\theta}^{(t-1)}$, 
	 from which the update loss $\ell\s{t-1}$ and  its gradient $\triangledown_{\boldsymbol{\theta}\s{t-1}}\ell\s{t-1}$ are computed using ground truth labels. Next, the element-wise stack \nabc{$\mathcal{I}^{(t-1)}$} consisting of previous learning rates, current parameters, current update loss and its gradient are input into a coordinate-wise LSTM \nabc{$\mathcal{O}$} to generate the adaptive learning rate \nabc{$\lambda^{(t-1)}$} as in (\ref{eq:2}). The model is then updated using one gradient descent step 
	\nnabc{(denoted by $\circleddash$)} as in (\ref{eq:3}). Finally, we apply the updated model \nabc{$\theta^{(t)}$} on a randomly selected future frame to get a meta loss for minimization as in (\ref{eq:5}). 
	}
	\label{fig:1}
	\vspace{-3ex}
\end{figure*}

\section{Proposed Algorithm}
Our tracker consists of two main modules: 1) a tracking model that is resizable to adapt to shape changes;  and 2) a neural optimizer that is in charge of model updating. The tracking model contains two branches where the response generation branch determines the presence of target by predicting a confidence score map and the bounding box regression branch estimates the precise box of the target by regressing coordinate shifts from the box anchors mounted on the sliding-window positions. The offline learned neural optimizer is 
\nabc{trained using a meta-learning framework}
 to online update the tracking model in order to adapt to appearance variations. Note both response generation and bounding box regression are built on the feature map computed from the backbone CNN network. The whole framework is briefly illustrated on Fig.~\ref{fig:1}

\subsection{Resizable Tracking Model}

\nyty{Trackers like correlation filter \cite{Henriques2015} and MetaTracker \cite{Park2018} initialize a convolutional filter based on the size of object in the first frame and keep its size fixed during subsequent frames. This setting is based on the assumption that the aspect ratio of object does not change during tracking, which is however often violated. Therefore, dynamically adapting the convolutional filter to the object shape variations is desirable, which means the number of filter parameters may vary between frames in the video and among different sequences.}
However, this complicates the design of neural optimizer when using separate learning rates for each filter parameter.
\nabc{To simplify the meta-learning framework and to better allow for per-parameter learning rates, we define fixed-shape convolutional filters, which are warped to the desired target size using bilinear interpolation before convolving with the feature map. In subsequent frames, the recurrent optimizer updates the fixed-shape tracking model.} \nyty{Note that MetaTracker \cite{Park2018} also resizes filters to the object size for model initialization, However, instead of dynamically adapting the convolutional filters to the object size for subsequent frames, MetaTracker keep the same shape as the initial filter during the following frames}

Specifically, tracking model $\boldsymbol{\theta}$ contains two parts, i.e. correlation filter $\boldsymbol{\theta}_{cf}$ and bounding box regression filter $\boldsymbol{\theta}_{reg}$. They are both warped 
to adapt to the shape variation of target,
\begin{align}
\boldsymbol{\theta} &= [\boldsymbol{\theta}_{cf}, \boldsymbol{\theta}_{reg}],\\
\tilde{\boldsymbol{\theta}}_{cf} &= \mathcal{W}(\boldsymbol{\theta}_{cf}, \phi), \label{eq:00} \\
\tilde{\boldsymbol{\theta}}_{reg} &= \mathcal{W}(\boldsymbol{\theta}_{reg}, \phi),
 \label{eq:0}
\end{align}
where $\mathcal{W}$ resizes the convolutional filter to 
size $\phi = (f_r, f_c)$ using bilinear interpolation. 
The filter size is computed 
from  the width and height $(w, h)$ of the object in the \nnabc{previous} image patch
\nnabc{(and for symmetry the filter size is odd)},
\begin{align}
f_r =  \ceil{\frac{\rho h}{c}} - \ceil{\frac{ \rho h}{c}}\hspace{-8pt}\mod 2 + 1, \label{eq:04}\\
f_c = \ceil{\frac{\rho w}{c}} -\ceil{\frac{\rho w}{c}}\hspace{-8pt}\mod 2 + 1
\label{eq:05}
\end{align}
where $\rho$ is the scale factor to enlarge the filter size to cover some context information, and $c$ is the stride of feature map, and $\ceil{~}$ means ceiling. 
%
Because of the resizable filters, there is no need to enumerate different aspect ratios and scales of anchor boxes when performing bounding box regression. We only use one sized anchor on each spatial location whose size is corresponding to the shape of regression filter, 
\begin{align}
(a_w, a_h) = (f_c, f_r) / \rho,
\end{align}
This saves regression filter parameters and achieves faster speed.  Note that we update the filter size and its corresponding anchor box every $\tau$ frames, i.e. just before every model updating,  during both offline training and testing/tracking phases. Through this modification, we are able to initialize the tracking model with $\boldsymbol{\theta}\s{0}$ and recurrently optimize it in subsequent frames without worrying about the shape changes of the tracked object.

\subsection{Recurrent Model Optimization} \label{sec:3.1}
Traditional optimization methods have the problem of slow convergence due to small learning rates and limited training samples, while simply increasing the learning rate has the risk of the training loss wandering wildly. 
Instead, we design a recurrent neural optimizer, which is trained to converge the model to a good solution in a few gradient steps\footnote{We only use one gradient step in our experiment, while considering multiple steps is straightforward.}, to update the tracking model. Our key idea is based on the assumption that \emph{the best optimizer should be able to update the model to generalize well on future frames.} During the \emph{offline} training stage, we perform a one step gradient update on the tracking model using our recurrent neural optimizer and then minimize its loss
\nabc{on future frames}. 
Once the \emph{offline} learning stage is finished, we use this learned neural optimizer to recurrently update the tracking model to adapt to appearance variations. The optimizer trained in this way will be able to quickly converge the model update to generalize well for future frame tracking.

We denote the response generation network as $\mathcal{G}(F; \boldsymbol{\theta}_{cf}, \phi)$ and the bounding box regression network as $\mathcal{R}(F; \boldsymbol{\theta}_{reg}, \phi)$, where $F$ is the feature map input and $\boldsymbol{\theta}$ are the parameters. The tracking loss consists of two parts: response loss and regression loss,
\begin{align}
\nonumber  L(F, M, B;  \boldsymbol{\theta}, \phi) = ~& \lVert \mathcal{G}(F; \boldsymbol{\theta}_{cf}, \phi)- M  \rVert^2 + \\
& \lVert \mathcal{R}(F; \boldsymbol{\theta}_{reg}, \phi)- B  \rVert_{s} \label{eq:1}
\end{align}
where the first term is the L2 loss and the second term is the smooth L1 loss \cite{Ren2015}, and $B$ represents the 
ground truth box. Note we adopt parameterization of bounding box coordinates as in \cite{Ren2015}. $M$ is the corresponding label map which is built using a 2D Gaussian function \nabc{and the ground-truth object location $(x_0,y_0)$ and size $(w,h)$,}
\begin{align}
M(x,y)=\exp{\left(-\alpha\left(\tfrac{(x-x_0)^2}{\sigma_x^2}+\tfrac{(y-y_0)^2}{\sigma_y^2}\right )\right)}
\end{align}
where $(\sigma_x, \sigma_y)=(w/c, h/c)$, and $\alpha$ controls the shape of the response map. 
\ytyn{Note that we use past predictions as pseudo-labels when performing model updating during testing. We only use the ground truth during offline training.}

A typical tracking process updates the tracking model using historical training examples and then tests this updated model on the following frames until the next update.  We simulate this scenario in a meta-learning paradigm by recurrently optimizing the tracking model, \abc{and then testing it on a future frame}. 
Specifically, the tracking network is updated by
\begin{align}
\boldsymbol{\theta}\s{t} &= \boldsymbol{\theta}\s{t-1} - \boldsymbol{\lambda}\s{t-1}\odot \triangledown_{\boldsymbol{\theta}\s{t-1}} \ell\s{t-1},  \label{eq:3}
\end{align}
where $\boldsymbol{\lambda}\s{t-1}$ is a fully element-wise learning rate that has the same dimension as the tracking model parameters $\boldsymbol{\theta}\s{t-1}$, and $\odot$ is element-wise multiplication.  The learning rate is recurrently generated 
\nabc{using an LSTM with input consisting of the \abc{previous learning rate $\boldsymbol{\lambda}\s{t-2}$}, the current parameters $\boldsymbol{\theta}\s{t-1}$, the current update 
loss $\ell\s{t-1}$ and its gradient $\triangledown_{\boldsymbol{\theta}\s{t-1}}\ell\s{t-1}$,}
\begin{align}
\mathcal{I}\s{t-1}&=[\boldsymbol{\lambda}\s{t-2}, \triangledown_{\boldsymbol{\theta}\s{t-1}}\ell\s{t-1}, \boldsymbol{\theta}\s{t-1}, \ell\s{t-1}], \\
\boldsymbol{\lambda}\s{t-1} &= \sigma(\mathcal{O}(\mathcal{I}\s{t-1}; \boldsymbol{\omega})),  \label{eq:2}
\end{align}
where
$\mathcal{O}(\cdot; \boldsymbol{\omega})$ is a coordinate-wise LSTM  \cite{Andrychowicz2016} parameterized by $\boldsymbol{\omega}$, which shares the parameters across all dimensions of input, and $\sigma$ is the sigmoid function to bound the predicted learning rate. The LSTM input $\mathcal{I}\s{t-1}$ is formed by element-wise \nabc{stacking the 4 sub-inputs}
along a new axis\footnote{We therefore get an $|\boldsymbol{\theta}|\times 4$ matrix, where $|\boldsymbol{\theta}|$ is the number of parameters in $\boldsymbol{\theta}$. Note that the current update loss $\ell\s{t-1}$  is broadcasted to have compatible shape with other vectors. 
	To better understand this process, we can treat the input of LSTM as a mini-batch of vectors where $|\boldsymbol{\theta}|$ is the batch size and 4 is the dimension of the input vector.}.  The current \textit{update loss} $\ell\s{t-1}$ is computed from a mini-batch of $n$ updating samples, 
\begin{align}
\ell\s{t-1}=\frac{1}{n}\sum_{j=1}^{n}L(F_j, M_j, B_j; \boldsymbol{\theta}\s{t-1}, \phi\s{t-1}), \label{eq:4}
\end{align}
\abcn{where the updating samples $(F_j, M_j, B_j)$}
\yty{are collected from the previous $\tau$ frames, where $\tau$ is the frame interval \abc{between model updates during online tracking.} }
Finally, we test the newly updated model $\boldsymbol{\theta}\s{t}$ on a randomly selected future \abcn{frame}\footnote{We found that using more than one future frame does not improve performance but costs more time during the off-line training phase.} 
and obtain the {\em meta loss}, 
\begin{align}
\mathcal{L}\s{t} =L(F\s{t+\delta}, M\s{t+\delta}, B\s{t+\delta}; \boldsymbol{\theta}\s{t}, \phi\s{t-1}),
\label{eq:5} 
\end{align}
where $\delta$ is randomly selected within $[0, \tau-1]$. 

\begin{algorithm}[tb]   
	\caption{Offline training of our framework}
	\label{algo:1}   
	\begin{algorithmic}[1] 
		\REQUIRE ~~\\
		$p(\mathcal{V})$: distribution over training videos.\\    
		\ENSURE ~~ \\
		$\boldsymbol{\boldsymbol{\theta}}\s{0}, \boldsymbol{\boldsymbol{\lambda}}\s{0}$ : initial tracking model and learning rates.\\
		$\boldsymbol{\omega}$: recurrent neural optimizer.\\
		
		\STATE Initialize all network parameters.
		\WHILE{not done}
		\STATE Draw a mini-batch of videos: $\mathcal{V}_i\sim p(\mathcal{V})$
		\FOR{all $\mathcal{V}_i$}
		\STATE Compute $\boldsymbol{\boldsymbol{\theta}}\s{1} \leftarrow \boldsymbol{\theta}\s{0} - \boldsymbol{\lambda}\s{0}\odot \triangledown_{\boldsymbol{\theta}\s{0}} \ell\s{0}$  in \refeqn{eq:3}.
		\STATE Compute meta loss $\mathcal{L}\s{1} $ using \refeqn{eq:5}.
		\FOR{$t=\{1+\tau, 1+2\tau, \cdots, 1+(T-1)\tau\}$}
		\STATE Compute adaptive learning rate $\boldsymbol{\lambda}\s{t-1}$ using neural optimizer $\mathcal{O}$ as in \refeqn{eq:2}.
		\STATE Compute updated model $\boldsymbol{\theta}\s{t}$ using \refeqn{eq:3}.
		\STATE Compute meta loss $\mathcal{L}\s{t}$ using \refeqn{eq:5}.
		\ENDFOR
		\ENDFOR
		\STATE Compute averaged meta loss $\bar{\mathcal{L}}$ using \refeqn{eq:6}.
		\STATE Update $\boldsymbol{\boldsymbol{\theta}}\s{0}, \boldsymbol{\boldsymbol{\lambda}}\s{0}, \boldsymbol{\omega}$ by computing gradient of $\bar{\mathcal{L}}$.
		\ENDWHILE
	\end{algorithmic}  
\end{algorithm}

During offline training stage, we perform the aforementioned procedure on a mini-batch of videos and get an averaged meta loss for optimizing \abcn{the neural optimizer},
\begin{align}
\bar{\mathcal{L}} = \frac{1}{NT}\sum_{i=1}^{N}\sum_{t=1}^{T}\mathcal{L}_{\mathcal{V}_i}\s{t},
\label{eq:6} 
\end{align}
where $N$ is the batch size and $T$ is the number of model updates, and $\mathcal{V}_i\sim p(\mathcal{V})$ is a video clip sampled from the training set. It should be noted that the initial tracking model parameter $\boldsymbol{\theta}\s{0}$ and initial learning rate $\boldsymbol{\lambda}\s{0}$ are also trainable variables, which are jointly learned with the neural optimizer $\mathcal{O}$. By minimizing the averaged meta loss $\bar{\mathcal{L}}$, we aim to train a neural optimizer that can update the tracking model to generalize well on subsequent frames, as well as to  learn a beneficial initialization of the tracking model, which is broadly applicable to different tasks (i.e.~videos). The overall training process is detailed in Algorithm \ref{algo:1}.

\subsection{Random Filter Scaling}
\begin{figure}[t]
	\centering
	\includegraphics[width=0.93\linewidth]{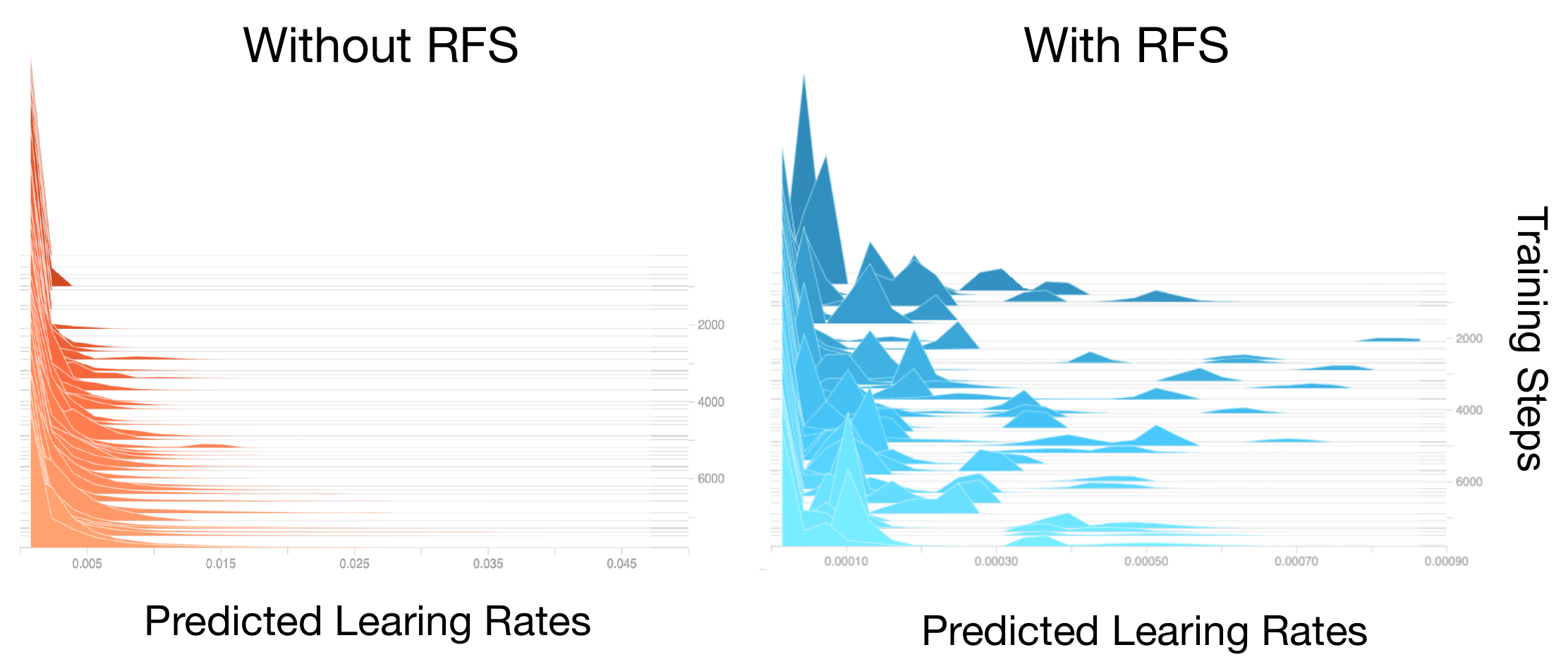}		
	\vspace{-2ex}
	\caption{Histogram of predicted learning rates during offline training}
	\label{fig:3}
	\vspace{-3ex}
\end{figure}

Neural optimizers have difficulty to generalize well on new tasks due to overfitting as discussed in \cite{Lv2017, Wichrowska2017,Andrychowicz2016}. By analyzing the learned behavior of the neural optimizer, we found that our preliminary trained optimizer will predict similar learning rates (see Fig.~\ref{fig:3} left).  We attribute this to overfitting to 
network inputs \nnabc{with similar magnitude scales}.
%
The following simple example illustrates the overfitting problem. Suppose the objective function\footnote{Our actual loss function in (\ref{eq:1}) includes an L2 loss and smooth L1 loss; for simplicity here we consider a simple linear model with a L2 loss.} that the neural optimizer minimizes is $g(\boldsymbol{\theta}) = \| \boldsymbol{x}\boldsymbol{\theta}-y \|^2$.
The optimal element-wise learning rate is $1/2\boldsymbol{x}^2$ since we can achieve the lowest loss of $0$ in one gradient descent step 
$\boldsymbol{\theta}\s{t+1}=\boldsymbol{\theta}\s{t}-1/2\boldsymbol{x}^2 \triangledown g(\boldsymbol{\theta}\s{t})=y/\boldsymbol{x}$. 
\nabc{Note that the optimal learning rate depends on the network input $\boldsymbol{x}$, and thus the learned neural optimizer is prone to overfitting if it does not see enough magnitude variations of $\boldsymbol{x}$.}
To address this problem, we multiply the tracking model $\boldsymbol{\theta}$ with a randomly sampled vector $\boldsymbol{\epsilon}$, which has the same dimension as $\boldsymbol{\theta}$ \cite{Lv2017} during each iteration of offline training,
\begin{align}
\boldsymbol{\epsilon} &\sim \exp(\mathcal{U}(-\kappa, \kappa)),  \label{eq:16}
\quad
\boldsymbol{\theta}_{\boldsymbol{\epsilon}} = \boldsymbol{\epsilon} \odot \boldsymbol{\theta}, 
\end{align}
where $\mathcal{U}(-\kappa, \kappa)$ is a uniform distribution in the interval $[-\kappa, \kappa]$, and $\kappa$ is the range factor to control the scale extent. The objective function is then modified as 
$g_{\boldsymbol{\epsilon}}(\boldsymbol{\theta}) = g(\frac{\boldsymbol{\theta}}{\boldsymbol{\epsilon}})$.
In this way, the network input $\boldsymbol{x}$ is indirectly scaled without modifying the training samples $(\boldsymbol{x}, y)$ in practice.  Thus, the learned neural optimizer is forced to predict adaptive learning rates  for inputs with different \nnabc{magnitudes} 
 (see Fig.~\ref{fig:3} right), rather than to produce similar learning rates for similar scaled inputs, which improves its generalization ability.  

\section{Online Tracking via Proposed Framework}
\nabc{The offline training produces a}
neural optimizer $\mathcal{O}$, initial tracking model $\boldsymbol{\theta}\s{0}$
and learning rate $\boldsymbol{\lambda}\s{0}$, 
which we then use to perform online tracking. The overall process is similar to offline training except that we do not compute the meta loss \abcn{or its gradient}. 

\textbf{Model Initialization.}~Given the first frame, the initial image patch is cropped \nabc{and} centered on the provided groundtruth bounding box. We then augment the initial image patch to 9 samples by stretching the object to different aspect ratios  and scales. Specifically, the target is stretched by $[swr, sh/r]$, where $(w, h)$ are the initial object width and height, and $(s, r)$ are scale and aspect ratio factors. 
Then we use these examples, as well as the offline learned $\boldsymbol{\theta\s{0}}$ and $\boldsymbol{\lambda\s{0}}$, to perform one-step gradient update to build an initial model $\boldsymbol{\theta\s{1}}$ as in \refeqn{eq:3}.

\textbf{Bounding Box Estimation.}~We estimate the object bounding box by first finding the presence of target through the response generation, and then predict accurate box by bounding box regression. 
We employ the penalty strategy used in \cite{Li2018} on the generated response to suppress \nyty{estimated boxes} with large changes in scale and aspect ratio.
 In addition, we also multiply the response map by a Gaussian-like motion map to suppress large movement. The bounding box computed by the anchor that corresponds to the maximum score of response map is the final prediction. To smoothen the results, we linearly interpolate this estimated object size with the previous one. 
\nabc{We denote our neural-optimized tracking model with bounding box regression as \textbf{ROAM++}. 
We also design a baseline variant of our tracker, which uses multi-scale search to the estimate object box instead of bounding box regression, and denote it as \textbf{ROAM}.}

\textbf{Model Updating.}~We update the model every $\tau$ frame. 
Although offline training uses the previous $\tau$ frames to perform a one-step gradient update of the model, in practice, we find that using more than one step could further improve the performance during tracking (see Sec.~\ref{sec:6.1}). Therefore, we adopt two-\abc{step} 
gradient update using the previous $2\tau$ frames in our experiments.

\section{Implementation Details} \label{sec:5}
\textbf{Patch Cropping.}~Given a bounding box of the object $(x_0, y_0, w, h)$, the ROI of the image patch has the same center \abc{$(x_0,y_0)$} 
and takes a larger size $S=S_w = S_h = \gamma \sqrt{w h}$, where $\gamma$ is \nnabc{the ROI} scale factor. Then, the ROI is resized to a fixed size $\mathcal{S}\times \mathcal{S}$ for batch processing.

\textbf{Network Structure.}~We use the first 12 convolution layers of the pretrained VGG-16 \cite{Simonyan2015} as the feature extractor. The top max-pooling layers are removed to increase the spatial resolution of the feature map. Both the response generation network $\mathcal{C}$ and bounding box regression network consists of two convolutional layers with a dimension reduction layer of 512$\times$64$\times$1$\times$1 \nabc{(in-channels, out-channels, height, width)} 
as the first layer, and either a correlation layer of 64$\times$1$\times$21$\times$21 or a regression layer of 64$\times$4$\times$21$\times$21 as the second layer respectively. We use two stacked LSTM layers with 20 hidden units for the neural optimizer $\mathcal{O}$.

The ROI scale factor is $\gamma = 5$ and the search size is $\mathcal{S}=281$. \nyty{The scale factor of filter size is $\rho=1.5$.} The response generation uses $\alpha=20$ and the feature stride of the CNN feature extractor is $c=4$. The scale and aspect ratio factors $s, r$ used for initial image patch augmentation are selected from $\{0.8, 1, 1.2\}$, generating a combination of 9 pairs of $(s, r)$. The range factors used in RFS are $\kappa_{cf}=1.6, \kappa_{reg}=1.3$ \footnote{We use different range factor $\kappa$ for $\boldsymbol{\theta}_{cf}$ and $\boldsymbol{\theta}_{reg}$ due to different magnitude of parameters in the two branches}.

\textbf{Training Details.} We use ADAM \cite{kingma2014adam} optimization with a mini-batch of 16 video clips of length $31$ on 4 GPUs (4 videos per GPU) to train our framework. 
 \nyty{We use the training splits of} ImageNet VID \cite{Krizhevsky2012}, TrackingNet \cite{Muller2018}, LaSOT \cite{Fan2019-1}, GOT10k \cite{Huang2019}, ImageNet DET \cite{Krizhevsky2012} and COCO \cite{lin2014microsoft} for training.  During training, we randomly extract a continuous sequence clip for video datasets, and repeat the same still image to form a video clip for image datasets. Note we randomly augment all frames in a training clip by slightly stretching and scaling the images. We use a learning rate of 1e-6 for the initial regression parameters $\boldsymbol{\theta\s{0}}$ and the initial learning rate $\boldsymbol{\lambda\s{0}}$. For the recurrent neural optimizer $\mathcal{O}$, we use a learning rate of 1e-3. Both learning rates are multiplied by 0.5 every 5 epochs. We implement our tracker in Python with the PyTorch toolbox \cite{paszke2017automatic}, and conduct the experiments on a computer with an NVIDIA RTX2080 GPU and Intel(R) Core(TM) i9 CPU @ 3.6 GHz. \nyty{Our tracker ROAM and ROAM++ run at 
 13 FPS and 20 FPS respectively. (See supplementary for detailed speed comparison)}

\section{Experiments}
We evaluate our trackers on six benchmarks: OTB-2015 \cite{Wu2015}, VOT-2016 \cite{Kristan2016}, VOT-2017 \cite{Kristan2017}, LaSOT \cite{Fan2019-1}, GOT-10k \cite{Huang2019} and TrackingNet \cite{Muller2018}.

\subsection{Comparison Results with State-of-the-art}

We compare our ROAM and ROAM++ with recent response generation based trackers including MetaTracker \cite{Park2018}, DSLT \cite{Lu2018}, MemTrack \cite{Yang2018},  CREST \cite{Song2017}, SiamFC \cite{Bertinetto2016}, CCOT \cite{Danelljan2016}, ECO \cite{Danelljan2016-1}, Staple \cite{Bertinetto2016-1}, as well as recent state-of-the-art trackers including SiamRPN\cite{Li2018}, DaSiamRPN \cite{Zhu2018}, SiamRPN+ \cite{Zhang2019} and C-RPN \cite{Fan2019} on both OTB and VOT datasets. 
\abcn{For the methods using SGD updates, the number of SGD steps followed their implementations.}



\begin{figure}[t]
	\centering
	\includegraphics[width=\linewidth]{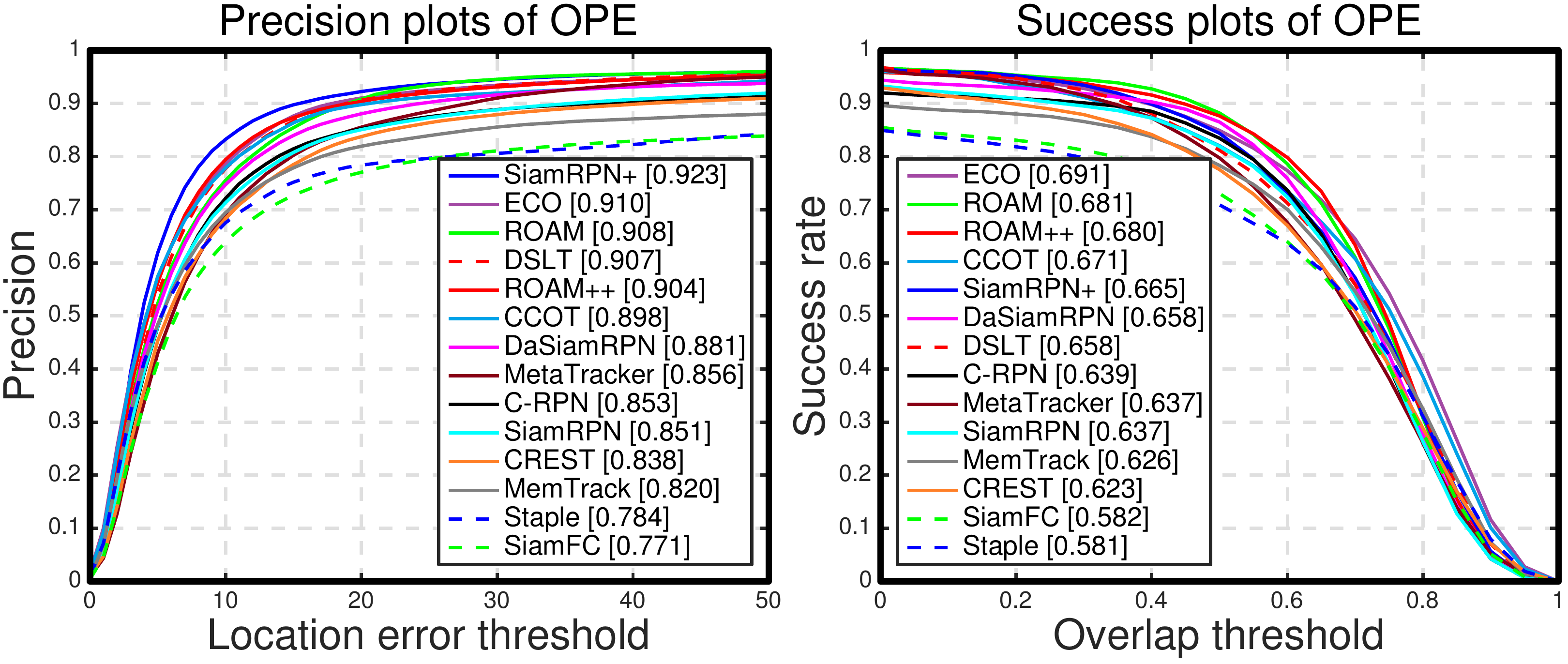}
			\vspace{-4ex}
	\caption{Precision and success plot on OTB-2015.}
	\label{fig:4}
	\vspace{-2ex}
\end{figure}

\begin{table}[]
	\small
	\begin{center}
		\bgroup
		\def\arraystretch{1.0}
		\scalebox{0.9}{\begin{tabular}{@{}cccc|ccc@{}}
				\toprule
				& \multicolumn{3}{c|}{VOT-2016} & \multicolumn{3}{c}{VOT-2017}  \\ \cline{2-7}
				& EAO($\uparrow$) & A($\uparrow$) & R($\downarrow$)   &   EAO($\uparrow$)  & A($\uparrow$)   & R($\downarrow$)           \\ \hline
				\textbf{ROAM++} & \first{0.441}& \second{0.599} &  \first{0.174} & \first{0.380} & \second{0.543}  & \first{0.195} \\
				\textbf{ROAM} & \third{0.384}& 0.556 &  \second{0.183} & \second{0.331} & 0.505  & \second{0.226} \\
				MetaTracker &0.317 &0.519 & - & - & - & -\\  \hline
				{DaSiamRPN} & \second{0.411} & \first{0.61} & \third{0.22}  & \third{0.326} & \first{0.56} & 0.34 \\
				{SiamRPN+} & {0.37} &{0.58} & 0.24 & {0.30} &\third{0.52} & 0.41 \\
				{C-RPN} & 0.363 & \third{0.594} & - &  {0.289} & - & -\\
				{SiamRPN} & 0.344 & 0.56 & 0.26 & 0.244 & 0.49 & 0.46 \\ \hline
				{ECO} &0.375 & 0.55 & {0.20} & 0.280 & 0.48 & \third{0.27} \\
				DSLT & 0.343 & 0.545 & 0.219 & - & - &- \\
				{CCOT} & {0.331} & 0.54  & 0.24 & 0.267 & 0.49 & 0.32 \\
				Staple & 0.295 & 0.54 & 0.38 & 0.169 & 0.52 & 0.69 \\
				CREST &0.283 & 0.51 & 0.25 & - & - & - \\
				MemTrack & 0.272 & 0.531  & 0.373 & 0.248 & {0.524} & 0.357 \\
				SiamFC & 0.235 & 0.532 & 0.461 & 0.188 & 0.502 & 0.585 \\ \bottomrule
		\end{tabular}}
		\egroup
	\end{center}
			\vspace{-4ex}
	\caption{Results on VOT-2016/2017. The evaluation metrics are expected average overlap (EAO), accuracy value (A), robustness value (R). The top 3 performing trackers are colored with \first{red}, \second{green} and \third{blue} respectively.}
	\vspace{-3ex}
	\label{tb:1}
\end{table}

\textbf{OTB.}~Fig.~\ref{fig:4} presents the experiment results on OTB-2015 dataset, which contains 100 sequences with 11 annotated video attributes.
Both our ROAM and ROAM++ achieve similar AUC compared with top performer ECO and outperform all other trackers. Specifically, our ROAM and ROAM++ surpass MetaTracker \cite{Park2018}, which is the baseline for meta learning trackers that uses traditional optimization method for model updating, by 6.9\% and 6.7\% on the success plot respectively, demonstrating the effectiveness of the proposed recurrent model optimization algorithm and resizable bounding box regression.  \nyty{In addition, both our ROAM and ROAM++ outperform the very recent meta learning based tracker MLT \cite{Choi2019} by a large margin (ROAM/ROAM++: 0.681/0.680 vs MLT: 0.611) under the AUC metric on OTB-2015. }

\textbf{VOT.} Table \ref{tb:1} shows the comparison performance on VOT-2016 and VOT-2017 datasets. Our ROAM++ achieves the best EAO on both VOT-2016 and VOT-2017. Specially, both our ROAM++ and ROAM shows superior performance on robustness value compared with RPN based trackers which have no model updating, demonstrating the effectiveness of our recurrent model optimization scheme. In addition, our ROAM++ and ROAM outperform the baseline MetaTracker \cite{Park2018} by  39.2\% and  21.1\% on EAO of VOT-2016, respectively.

\begin{figure}[t]
	\centering
	\includegraphics[width=\linewidth]{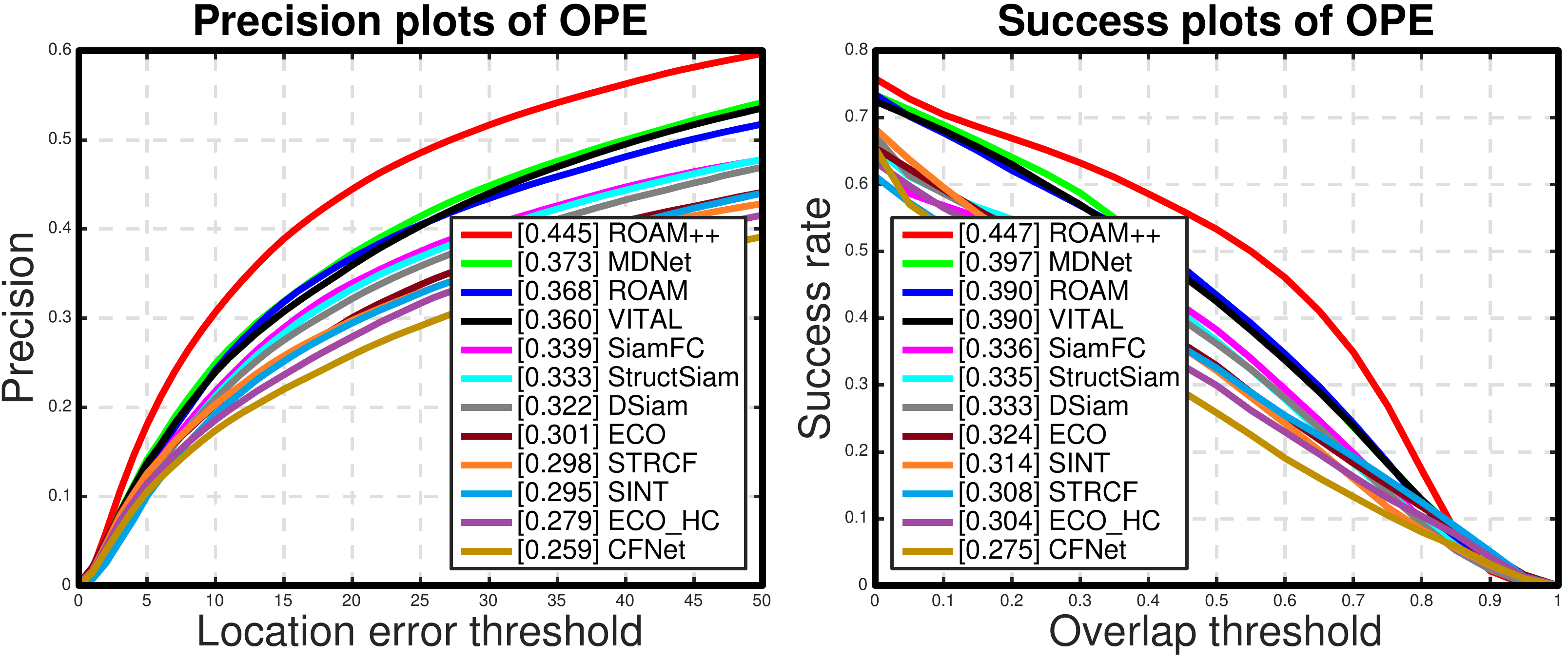}
		\vspace{-4ex}
	\caption{Precision and success plot on LaSOT test dataset}
	\label{fig:5}
	\vspace{-2ex}
\end{figure}

\textbf{LaSOT.} LaSOT  \cite{Fan2019-1} is a recently proposed large-scale tracking dataset. We evaluate our ROAM against the top-10 performing trackers of the benchmark  \cite{Fan2019-1},
 including MDNet \cite{Nam2016}, VITAL \cite{Song2018}, SiamFC \cite{Bertinetto2016}, StructSiam \cite{Zhang2018-1},   DSiam \cite{Guo2017}, SINT \cite{Guo2017}, ECO \cite{Danelljan2016-1}, STRCF \cite{li2018learning}, ECO\_HC \cite{Danelljan2016-1} and CFNet \cite{Valmadre2017}, on the testing split which consists of 280 videos. Fig.~\ref{fig:5} presents the comparison results of precision plot and success plot on LaSOT testset. Our ROAM++ achieves the best result compared with state-of-the-art trackers on the benchmark, outperforming the second best MDNet with an improvement of 19.3\% and 12.6\% on precision plot and success plot respectively. 

\textbf{GOT-10k.} GOT-10k \cite{Huang2019} is large highly-diverse dataset for object tracking that is also proposed recently. There is no overlap in object categories between the training set and test set, which follows the \textit{one-shot learning setting} \cite{Finn2017}. Therefore, using external training data is strictly forbidden when testing trackers on their online server. Following their protocol, we train our ROAM by using only the training split of this dataset. 
Table \ref{tb:2} shows the detailed comparison results on GOT-10k test dataset. Both our ROAM++ and ROAM surpass other trackers with a large margin. In particular, our ROAM++ obtains an AO of 0.465, a $\text{SR}_{0.5}$ of 0.532 and a $\text{SR}_{0.75}$ of 0.236, outperforming SiamFCv2 with an improvement of  24.3\%,  31.7\% and 63.9\% respectively. 
\begin{table}[]
	\small
	\begin{center}
		\bgroup
		\def\arraystretch{1.0}\tabcolsep=2pt
		\scalebox{0.68}{
			\begin{tabular}{cccccccccc}
				\toprule
				& \tabincell{c}{MDNet \\\cite{Nam2016}} & \tabincell{c}{CF2\\\cite{Ma2015}} & \tabincell{c}{ECO \\\cite{Danelljan2016-1}} & \tabincell{c}{CCOT \\\cite{Danelljan2016}} & \tabincell{c}{GOTURN \\\cite{Held2016}} & \tabincell{c}{SiamFC \\\cite{Bertinetto2016}} & \tabincell{c}{SiamFCv2 \\\cite{Valmadre2017}} & \textbf{ROAM} & \textbf{ROAM++}\\ \midrule
				AO($\uparrow$) & 0.299 & 0.315 & 0.316 & 0.325 & 0.342 & {0.348} & \third{0.374} & \second{0.436} & \first{0.465} \\
				$\text{SR}_{0.5}$($\uparrow$) & 0.303 & 0.297 & 0.309 & 0.328 & {0.372} & 0.353 & \third{0.404} & \second{0.466} & \first{0.532} \\
				$\text{SR}_{0.75}$($\uparrow$) & 0.099 & 0.088 & 0.111 & 0.107 & {0.124} & 0.098 & \third{0.144} & \second{0.164} & \first{0.236} \\\bottomrule
			\end{tabular}
		}
		\egroup
	\end{center}
		\vspace{-4ex}
	\caption{Results on GOT-10k. The evaluation metrics include average overlap (AO), success rate at 0.5 overlap threshod. ($\text{SR}_{0.5}$ ), success rate at 0.75 overlap threshod. ($\text{SR}_{0.75}$ ). The top three performing trackers are colored with \first{red}, \second{green} and \third{blue} respectively.}
	\label{tb:2}
	\vspace{-4ex}
\end{table}

\textbf{TrackingNet.} TrackingNet \cite{Muller2018} provides more than 30K videos with around 14M dense bounding box annotations by filtering shorter video clips from Youtube-BB \cite{Brain2017}.  Table \ref{tb:3} presents the detailed comparison results on TrackingNet test dataset. Our ROAM++ surpasses other state-of-the-art tracking algorithms on all three evaluation metrics. In detail, our ROAM++ obtains an improvement of 10.6\%, 10.3\% and 6.9\% on AUC, precision and normalized precision respectively compared with top performing tracker MDNet on the benchmark.

\begin{table}[]
	\small
	\begin{center}
		\bgroup
		\def\arraystretch{1.0}\tabcolsep=2pt
		\scalebox{0.68}{
			\begin{tabular}{cccccccccc}
				\toprule
				&\tabincell{c}{Staple \\\cite{Bertinetto2016-1}}& \tabincell{c}{CSRDCF \\\cite{Lukezic2017}} & \tabincell{c}{ECOhc \\\cite{Danelljan2016-1}} & \tabincell{c}{ECO \\\cite{Danelljan2016-1}} & \tabincell{c}{SiamFC \\\cite{Bertinetto2016}} & \tabincell{c}{CFNet \\\cite{Valmadre2017}}& \tabincell{c}{MDNet \\\cite{Nam2016}} & \textbf{ROAM} & \textbf{ROAM++} \\ \midrule
				AUC($\uparrow$) & 0.528 & 0.534 & 0.541 & 0.554 & 0.571 & {0.578} & \third{0.606} & \second{0.620} & \first{0.670} \\
				Prec.($\uparrow$) & 0.470 & 0.480 & 0.478 & 0.492 & {0.533} & {0.533} & \second{0.565} & \third{0.547} & \first{0.623}\\
				Norm. Prec.($\uparrow$) & 0.603 & 0.622 & 0.608 & 0.618 & {0.663} & 0.654 & \second{0.705} & \third{0.695} & \first{0.754} \\\bottomrule
			\end{tabular}
		}
		\egroup
	\end{center}
		\vspace{-4ex}
	\caption{Results on TrackingNet. The evaluation metrics include area under curve (AUC) of success plot, Precision, Normalized Precison. The top three performing trackers are colored with \first{red}, \second{green} and \third{blue} respectively.}
	\label{tb:3}
	\vspace{-2ex}
\end{table}


\subsection{Ablation Study} \label{sec:6.1}
For a deeper analysis, 
we study our trackers from various aspects. Note that all these ablations are trained only on ImageNet VID dataset for simplicity. 

\begin{figure}[tb]
	\centering
	\includegraphics[width=\linewidth]{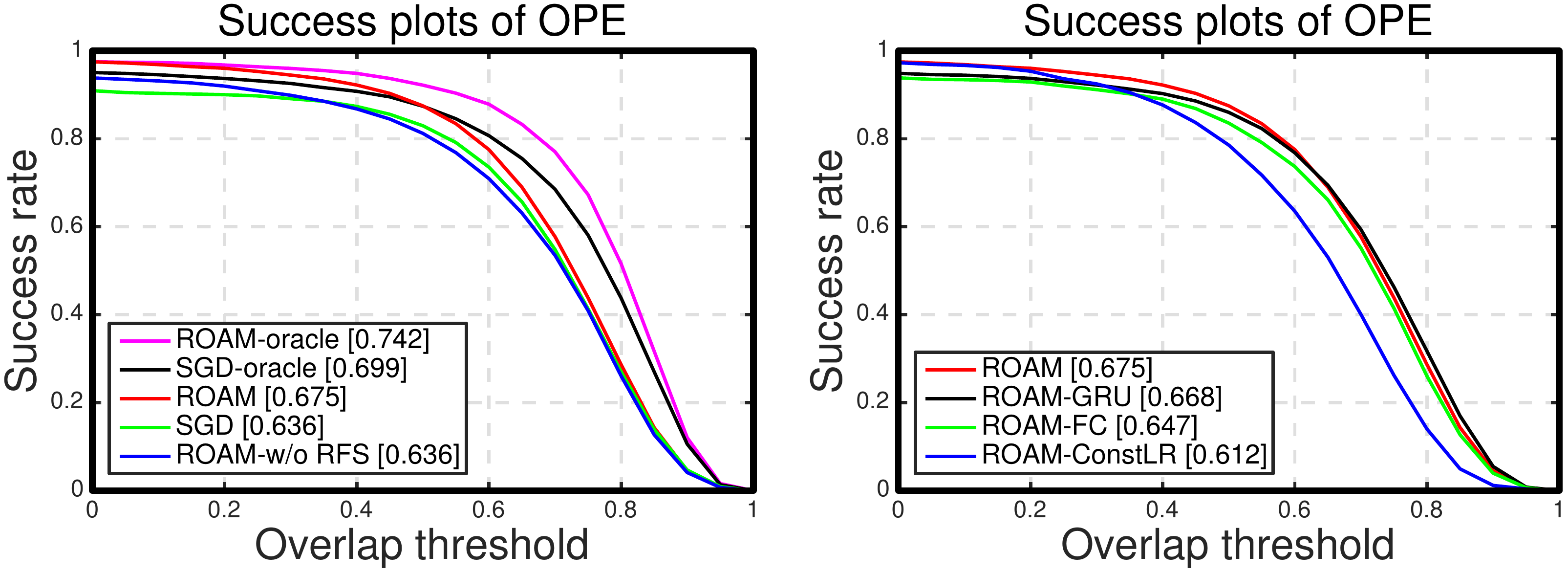}
	\vspace{-4ex}
	\caption{Ablation Study on OTB-2015 with different variants of ROAM. }
	\label{fig:6}
	\vspace{-3ex}
\end{figure} 

\textbf{Impact of Different Modules.}~To verify the effectiveness of different modules, we design four variants of our framework: 1) \textit{SGD}: replacing recurrent neural optimizer with traditional SGD  for model updating (\yty{using the same number of gradient steps as ROAM}); 
2) \textit{ROAM-w/o RFS}: training a recurrent neural optimizer without RFS;  3) \textit{SGD-Oracle}: using the ground-truth bounding boxes to build updating samples for SGD during the testing phase (\nyty{using the same number of gradient steps as ROAM}); 
4) \textit{ROAM-Oracle}: using the ground-truth bounding boxes to build updating samples for ROAM during the testing phase. The results are presented in Fig.~\ref{fig:6} (Left). ROAM gains about 6\% improvement on AUC compared with the baseline SGD, demonstrating the effectiveness of our recurrent model optimization method on model updating. Without RFS during offline training, the tracking performance drops substantially due to overfitting.  ROAM-Oracle performs better than SGD-Oracle, showing that our offline learned neural optimizer is more effective than traditional SGD method given the same updating samples. In addition, these two oracles (SGD-oracle and ROAM-oracle) achieve  higher AUC score compared with their normal versions, indicating that the tracking accuracy could be  boosted by improving the quality of the updating samples. 

\textbf{Architecture of Neural Optimizer.} To investigate more architectures of the neural optimizer, we presents three variants of our method: 1) \textit{ROAM-GRU}: using two stacked Gated Recurrent Unit (GRU) \cite{cho2014learning}  as our neural optimizer; 2) \textit{ROAM-FC}: using two linear fully-connected layers followed by tanh activation function as the neural optimizer; 3) \textit{ROAM-ConstLR}: using a learned {\em constant} element-wise learning rate for model optimization instead of the adaptively generated one.  Fig.~\ref{fig:6} (Right) presents the  results.
using ROAM-GRU decreases the AUC  slightly,  while
ROAM-FC has significantly lower AUC compared with ROAM, showing the importance of our recurrent structure. Moreover, the performance drop of ROAM-ConstLR verifies the necessity of using an adaptable learning rate for model updating.

\begin{figure}[tb]
	\centering
	\includegraphics[width=\linewidth]{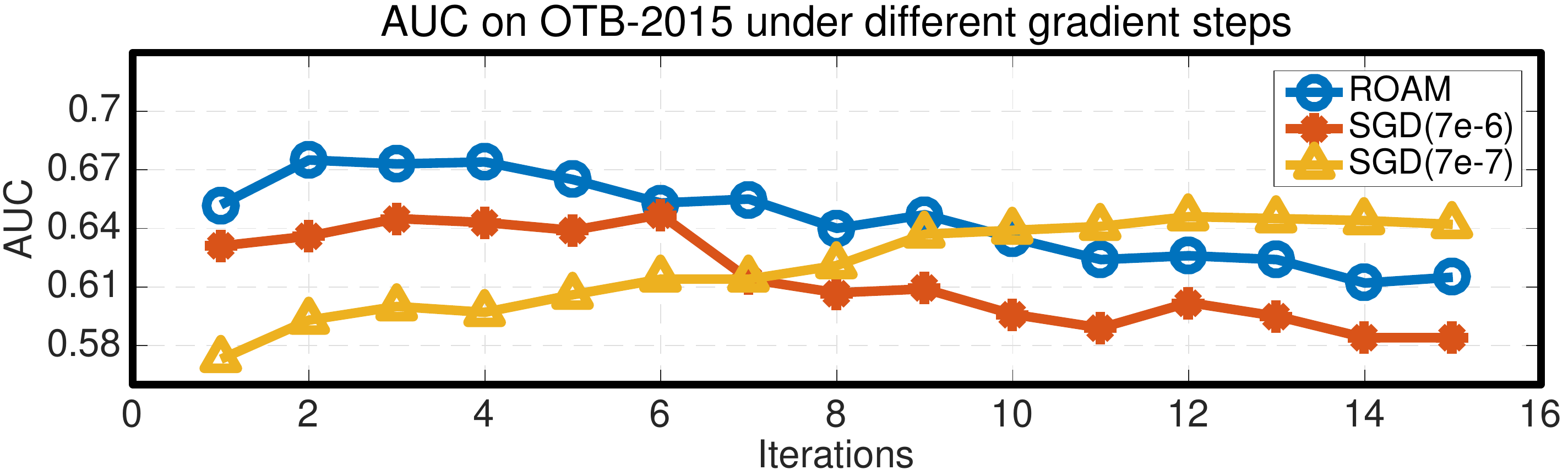}
		\vspace{-4ex}
	\caption{AUC vs. gradient steps on OTB-2015.}
	\label{fig:7}
	\vspace{-2ex}
\end{figure}

\textbf{More Steps in Updating.}~During offline training, we only perform one-step gradient decent \abcn{to optimize the updating loss.}
We investigate the effect of using more than one gradient step on tracking performance during the test phase  for both ROAM and SGD (see Fig.~\ref{fig:7}). Our method can be further improved with more than one step, but will gradually decrease when using too many steps. This is because our framework is not trained to perform so many steps during the offline stage. \yty{We also use two fixed learning rates for SGD, where the larger one is 7e-6 \footnote{MetaTracker uses this learning rate.} and the smaller one is 7e-7. Using a larger learning rate, SGD could reach its best performance much faster than using a smaller learning rate, while both have similar best AUC.
	Our ROAM consistently outperforms SGD(7e-6), showing the superiority of adaptive element-wise learning rates.} 
\abcnn{Furthermore, ROAM with 1-2 gradient steps outperforms SGD(7-e7) using a large number of steps, which shows the improved generalization of ROAM.}

\begin{figure}[tb]
	\centering
	\includegraphics[width=0.95\linewidth]{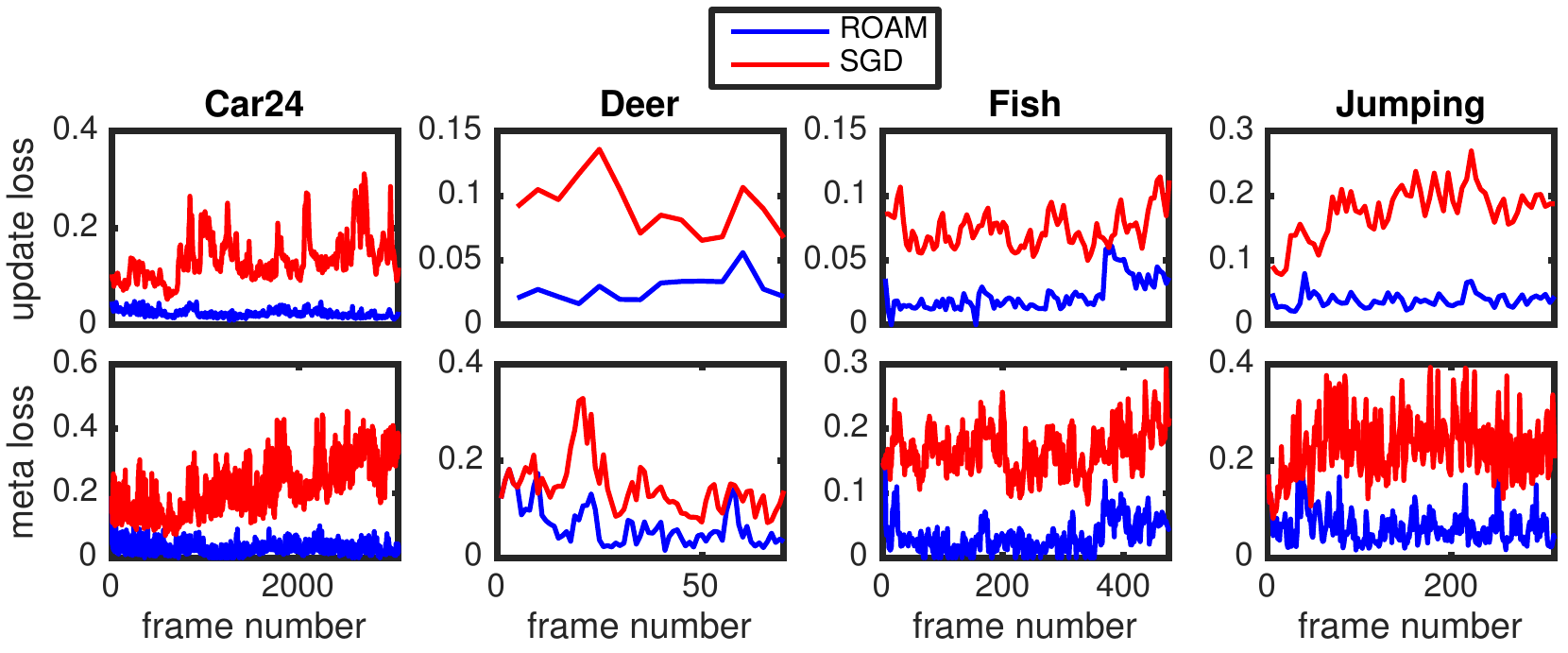}
	\vspace{-2ex}
	\caption{Update loss (top row) and meta loss (bottom row) comparison between ROAM and SGD.
	}
	\label{fig:8}
	\vspace{-3ex}
\end{figure}

\textbf{Update Loss and Meta Loss Comparison.}  To show the effectiveness of our neural optimizer, we compare the update loss and meta loss over time between ROAM and SGD after two gradient steps for a few videos of OTB-2015 in Fig.~\ref{fig:8} (see supplementary for more examples).  
Under the same number of gradient updates, our neural optimizer obtains lower loss compared with traditional SGD, demonstrating its faster converge and better generalization ability for model optimization.

\begin{figure}[t]
	\centering
	\includegraphics[width=\linewidth]{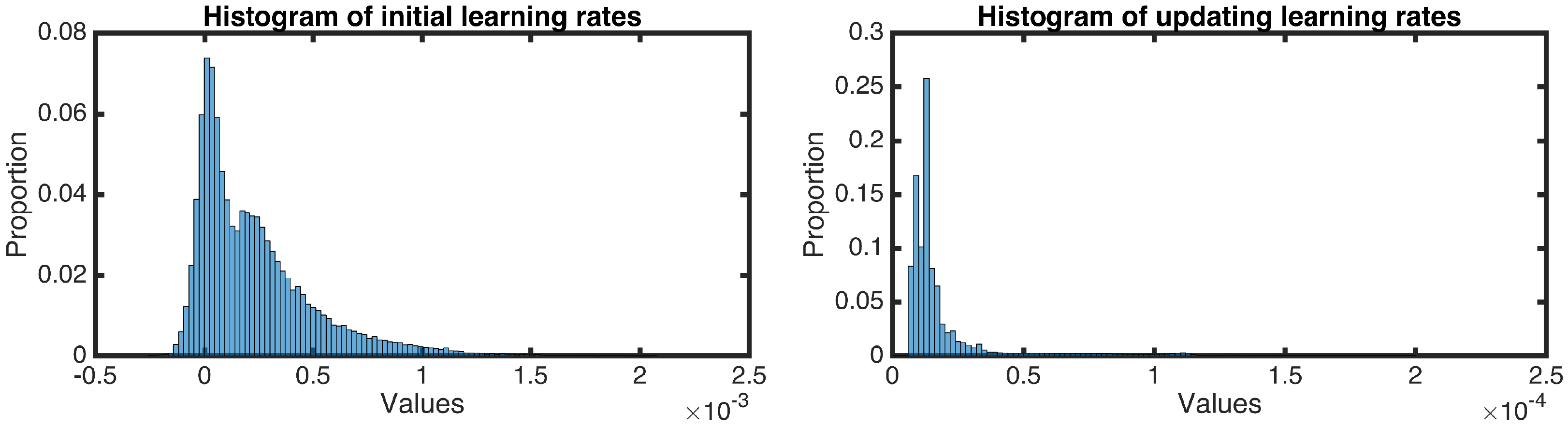}
			\vspace{-4ex}
	\caption{Histogram of initial learning rates and updating learning rates on OTB-2015.
}
	\label{fig:9}
	\vspace{-4ex}
\end{figure}

\textbf{Why does ROAM work?}~As discussed in \cite{Park2018}, directly using the learned initial learning rate $\boldsymbol{\lambda}\s{0}$ for  model optimization in subsequent frames could lead to divergence.  This is because the learning rates for model  initialization 
are relatively larger than the ones needed for subsequent frames, which therefore causes unstable model optimization. In particular, the initial 
model  $\boldsymbol{\theta}\s{0}$ is offline trained to be broadly applicable to different videos, which therefore needs relatively larger gradient step to adapt to a specific task, leading to a relative larger $\boldsymbol{\lambda}\s{0}$. For the subsequent frames, the appearance variations could be sometimes small or sometimes large, and thus the model optimization process needs an adaptive learning rate to handle different situations.  Fig.~\ref{fig:9} presents the histograms of initial learning rates and updating learning rates on OTB-2015. 
Most of updating learning rates are relatively small because usually there are only minor appearance variations between updates.
As is shown in Fig.~\ref{fig:4}, our ROAM, which performs model updating for subsequent frames with adaptive learning rates, obtains substantial performance gain compared with MetaTracker \cite{Park2018}, which uses a traditional SGD with a constant learning rate for model updating.   

\section{Conclusion}
In this paper, we propose a novel tracking model consisting of resziable response generator and bounding box regressor, where the first part generates a score map to indicate the presence of target at different spatial locations and the second module regresses bounding box coordinates shift to the anchors densely mounted on sliding window positions. 
To effectively update the tracking model for appearance adaption, we propose a recurrent model optimization method within a meta learning setting in a few gradient steps. Instead of producing an increment for parameter updating, which is prone to overfit due to different gradient scales, we recurrently generate adaptive learning rates for tracking model optimization using an LSTM. 
Extensive experiments on OTB, VOT, LaSOT, GOT-10k and TrackingNet datasets demonstrate superior performance compared with state-of-the-art tracking algorithms. 

\CUT{
	Besides training an effective optimizer for model updating, the quality of updating examples is also crucial to improve tracking performance. As is shown in Fig.~\ref{fig:2} (left), ROAM-oracle, which utilizes the ground truth bounding box to construct training examples for model updating, achieves tremendous accuracy gain, indicating that examples built using the predicted bounding boxes contains label noise. Therefore, finding a way to alleviate the adverse effect of noisy examples could be future work.
	\NOTE{isn't this true for any online updating model? If you use the true bbox to update the model you will always get better performance?. \yty{Yes, better quality updating examples will improve performance.}}
}



\noindent \textbf{Acknowledgement:} This work was supported by grants from the Research Grants Council of the Hong Kong Special Administrative Region, China (Project No. [T32-101/15-R] and CityU 11212518).

{\small
    \bibliographystyle{ieee_fullname}
	\bibliography{egbib}
}

\end{document}